\documentclass{article}
\usepackage{ijcai24}

\usepackage{amssymb}
\usepackage{times}
\usepackage{soul}
\usepackage{url}
\usepackage[hidelinks]{hyperref}
\usepackage[utf8]{inputenc}
\usepackage[small]{caption}
\usepackage{graphicx}
\usepackage{amsmath}
\usepackage{amsthm}
\usepackage{booktabs}
\usepackage{algorithm}
\usepackage{algorithmic}
\usepackage[switch]{lineno}

\title{Unified-IoU: For High-Quality Object Detection}

\author{
Xiangjie Luo$^1$
\and
Zhihao Cai$^{1,2*}$
\and
Bo Shao$^1$\And
Yingxun Wang$^{1,2}$\\
\affiliations
$^1$School of Automation Science and Electrical Engineering, Beihang University\\
$^2$Institute of Unmanned System, Beihang University\\
\emails
luoxiangjie@buaa.edu.cn,
czh@buaa.edu.cn,
ZB2103011@buaa.edu.cn,
wangyx@buaa.edu.cn
}

\begin{document}

\maketitle

\begin{abstract}
    Object detection is an important part in the field of computer vision, and the effect of object detection is directly determined by the regression accuracy of the prediction box. As the key to model training, IoU (Intersection over Union) greatly shows the difference between the current prediction box and the Ground Truth box. Subsequent researchers have continuously added more considerations to IoU, such as center distance, aspect ratio, and so on. However, there is an upper limit to just refining the geometric differences; And there is a potential connection between the new consideration index and the IoU itself, and the direct addition or subtraction between the two may lead to the problem of "over-consideration". Based on this, we propose a new IoU loss function, called Unified-IoU (UIoU), which is more concerned with the weight assignment between different quality prediction boxes. Specifically, the loss function dynamically shifts the model's attention from low-quality prediction boxes to high-quality prediction boxes in a novel way to enhance the model's detection performance on high-precision or intensive datasets and achieve a balance in training speed. Our proposed method achieves better performance on multiple datasets, especially at a high IoU threshold, UIoU has a more significant improvement effect compared with other improved IoU losses. Our code is publicly available at: https://github.com/lxj-drifter/UIOU\_files.
\end{abstract}

\section{Introduction}

Bounding box regression (BBR) module is an important part of target detection model~\cite{bib23,bib24,bib25,bib26,bib27,bib19,bib28}, a good and accurate bounding box regression function directly determines the accuracy of target location. If the positioning accuracy of the prediction box is poor, it is very easy to cause the false or missing detection of the target. Bounding box regression modules have been widely used in many advanced target detectors, including Mask R-CNN~\cite{bib1}, Cascade R-CNN~\cite{bib2}, YOLO~\cite{bib3} and so on. Therefore, designing a good BBR loss function is the key of object detection task. At present, BBR loss functions are mainly divided into two categories:

The first type of regression based BBR loss is defined as $\ell_2$ loss~\cite{bib4,bib5}, which divides the height and width of prediction box and bounding box into two parts respectively by given pixels, and the position information of each bounding box can be defined as a four-dimensional vector:%
\begin{equation}
	\resizebox{.41\linewidth}{!}{$
		\displaystyle
		x = (x_t, x_b, x_l, x_r)
		$}.
\end{equation}%
where $x_t$, $x_b$, $x_l$, $x_r$ can be understood as the distance between the pixel position and the top, bottom, left and right borders of the bounding box. The $\ell_2$ loss can be calculated as:%
\begin{equation}
	\resizebox{.51\linewidth}{!}{$
		\displaystyle
		\ell_2 \quad\quad loss = ||x_1 - x_2||^2_2
		$}.
\end{equation}%
where $x_1$ represents the four-dimensional vector of the prediction box, and $x_2$ represents the four-dimensional vector of the Ground Truth box. $\ell_2$ loss normalized a number of features, and the results are all between 0 and 1. However, the four variables in the four-dimensional vector are independent of each other, and the correlation of the boundary is ignored. Moreover, the method only considers the distance between the pixel and the boundary, and has no limit on the size of the bounding box, so the model has poor localization performance for small targets~\cite{bib6,bib8}.

The second type of BBR loss is IoU (Intersection over Union) loss~\cite{bib6}. The prediction box is defined as $P$, the Ground Truth box as $P^{gt}$, and the calculation formula of IoU loss is:%
\begin{equation}
	\resizebox{.51\linewidth}{!}{$
		\displaystyle
		IoU \quad\quad loss = 1 - \frac{P\cap P^{gt}}{P\cup P^{gt}}
		$}.
\end{equation}%

The above two types of BBR loss are standardized, but they are not sensitive to the scale of the bounding box, and only focus on the problem of the geometric scale, ignoring the quality and difficulty of the bounding box.

In this paper, we focus the attention of the model on the different quality anchor boxes, and no longer modify the geometric metric rules between the bounding boxes. Firstly, we propose a new dynamic weighting method for prediction box. By reducing the bounding box, we can enlarge the IoU loss of prediction box and GT box, which is equivalent to giving more weight to high-quality prediction box and achieving the effect of focusing the model on high-quality prediction box. Shrinking the bounding box has the opposite effect. Secondly, for high-quality object detection, we want to focus the model's attention on high-quality anchor boxes, but this will bring the problem of slow convergence. In order to balance the contradiction between the model's attention and the convergence rate, we designed a dynamic hyperparameter "ratio" to adjust the scaling ratio of the bounding box. Specifically, at the beginning of training, we enlarged the bounding box, which is equivalent to reducing the IoU loss of the high-quality prediction box, and the model focused on the low-quality anchor box, which can converge faster. At the later stage of training, we reduce the bounding box, which is equivalent to increasing the IoU loss of the high-quality prediction box, so that the model's attention is focused on the high-quality anchor box. Thirdly, inspired by "Focal Loss"~\cite{bib7}, we designed a dual attention mechanism for bounding box regression (BBR) loss to further optimize weight allocation. Finally, we combined all the above innovations to design a new IoU loss function, called Unified-IoU (UIoU), to achieve high-quality object detection. In order to verify our proposed method, we performed a comparison experiment with all known improvement methods (e.g. GIoU~\cite{bib8}, CIoU~\cite{bib9}, $\alpha$IoU~\cite{bib10}, EIoU~\cite{bib11}, SIoU~\cite{bib12}, WIoU~\cite{bib13}). Consistent and significant improvements across multiple datasets illustrate the potential of our new UIoU loss function.

Our contributions in this paper are as follows:
\begin{itemize}
	\item We design a novel method to dynamically assign weights to different quality anchor boxes during model training.
	\item Considering the contradiction between the model convergence speed and high-quality detection, a strategy of dynamically shifting the model's attention by using the hyperparameter "ratio" is proposed, which achieves better convergence speed and regression results than the original.
	\item Drawing on the idea of "Focal Loss", we design a dual attention mechanism for the box regression loss to further optimize the weight allocation of anchor boxes with different quality.
	\item Combined with the above innovation points, UIoU is designed as a new bounding box regression loss function. Extensive experiments are conducted to demonstrate the superiority of the proposed method, and the performance of various improvements is verified in the ablation experiment.
\end{itemize}

\section{Related Work}

\begin{figure}[h]%
	\centering
	\includegraphics[width=0.45\textwidth]{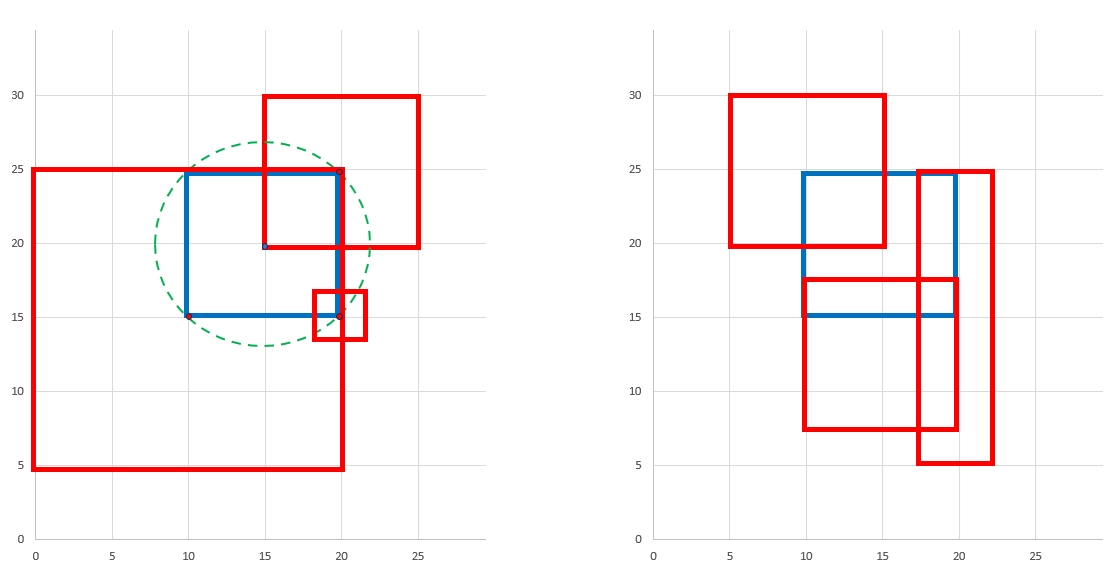}
	\caption{The existing weight allocation strategy: using center distance or IoU value as the weight allocation factor is obviously unacceptable (The three red boxes in the left image represent the prediction boxes, they have the same center distance as the blue GT box, but it is obviously unreasonable to give them the same weight; The three red boxes in the right image represent the predicted boxes that have the same IoU value as the blue GT box, and there is a large difference between them, it is also unreasonable to give the same weight)}\label{fig1}
\end{figure}

\subsection{IoU Loss in Geometric Metric}

IoU loss is the most effective loss function paradigm for current target detectors~\cite{bib29,bib30,bib31,bib32,bib33}. In the course of IoU research, a lot of work is focused on the geometric measurement of anchor box and GT box. GIoU~\cite{bib8} solves the problem of gradient disappearance caused by disjoint boxes of the original IoU, and introduces the minimum closed shape $C$, which can include both prediction box $P$ and Ground Truth box $P^{gt}$, and can also drive the prediction box towards the GT box when the two boxes are disjoint. However, GIoU does not solve the problem in the case of inclusion of two boxes, in which case GIoU degenerates into IoU. On the basis of GIoU, CIoU~\cite{bib9} adds the consideration of center distance and bounding box scale to make the target box regression more stable, but it adopts the aspect ratio to measure the difference between the two boxes. If the aspect ratio of the prediction box and the GT box is the same, then the penalty for this item is always 0, which is unreasonable. Based on the penalty item of CIoU, EIoU~\cite{bib11} dissolves the influence factors of the aspect ratio of the prediction box and the GT box, and calculates the width and height of the prediction box and the GT box respectively to solve the problems existing in CIoU. SIoU~\cite{bib12} takes into account the angle of the vector between the prediction box and the GT box, and redefines the angle penalty metric, which allows the prediction box to quickly drift to the nearest axis and then only need to return one coordinate ($X$ or $Y$), which effectively reduces the total number of degrees of freedom. However, SIoU requires multiple IoU thresholds to be set for calculation, and these thresholds need to be adjusted continuously with datasets and tasks, which makes evaluation difficult and results lack confidence.

Throughout the above improvements to the IoU loss, all of them are the continuous mining of the geometric metric of the prediction box and the GT box. Although they solve the problems brought by the IoU, they lead to excessive coupling between various indicators. For example, between the IoU itself and the center distance, when we keep other terms unchanged, reduce the center distance, the maximum probability of IoU will also change, which results in the ambiguity of the role of various indicators. At present, the optimization of IoU geometric measures is almost up to the limit, and we should focus on those examples that are more useful for bounding box regression.

\subsection{Effective Example Mining}

There is a large number of sample imbalance problems in BBR, and the number of low-quality prediction boxes is often much more than that of high-quality prediction boxes, which results in the excessive influence of low-quality prediction boxes in gradient update~\cite{bib19,bib20}. In this way, the trained model will produce large bounding box regression errors in target detection, especially when facing dense datasets, and it is easy to produce a lot of missing and false detection.

Different studies have different definitions of anchor box quality. It is common to use center distance or IoU value to define the quality of anchor box and weight factor, but sometimes this is wrong, as shown in Figure~\ref{fig1}. $\alpha$IoU~\cite{bib10} generalizes the existing IOU-based Loss to a new family of power IoU losses, and adjusts the weight of each index in IoU Loss through a single power parameter $\alpha$. WIoU~\cite{bib13} constructed a weight factor $R_{WIoU}$ based on center distance and minimum closed shape $C$, and a non-monotonic focusing coefficient $\beta$, to obtain a loss function with dynamic non-monotonic FM. However, these methods can only increase the gradient of high-quality prediction boxes, but can not suppress the abnormal gradient of low-quality prediction boxes~\cite{bib19,bib20}, and there are also problems of slow convergence of the model.

Focal Loss~\cite{bib7} interprets "effective example" from the perspective of "confidence". Instead of starting from IoU, it relies on the classification difficulty of prediction box. It believes that a large number of easy classification examples produce too much gradient, resulting in insufficient attention of the model to difficult classification target examples. However, it is only suitable for binary classification tasks, and the selection of hyperparameters $\gamma$ and $\alpha$ is difficult and the effect is unstable. SSD~\cite{bib21} prevents the predicted value of the network from approaching the negative sample by screening the negative sample, that is, the negative sample used to train the network is a subset of the extracted negative sample. OHEM~\cite{bib22} shifted its attention from focusing only on hard and negative samples to all hard samples, both positive and negative. These methods simply assign weights or filter out some unsatisfactory prediction boxes, which do not fully adapt to the needs of different quality anchor boxes in different training stages.

\section{Method}

\subsection{Motivation}

For single-stage detector YOLO, the current improvement of bounding box regression loss function (IoU Loss) mainly focuses on the deviation of the prediction box and the Ground Truth (GT) box, trying to obtain more accurate IoU value by quantifying all aspects of the difference between the two boxes, such as the distance of the center point, the aspect ratio, the angle between the center point line and the x-y axis, and so on. We believe that this is inefficient, and the relationship quantities are coupled to each other, and it is difficult to show that the impact is determined by a single indicator or a few indicators.

In some high demand for regression accuracy or dense scenes, high-quality target detection requires high-quality bounding box regression loss, we should not only focus on those low quality prediction boxes and blindly give them more weight. Different from the previous work, we pay more attention to those high-quality prediction boxes, and design a new dynamic weight allocation method called "Focal Box". Combined with the idea of "Focal Loss", we propose Unified-IoU as a new loss function.

\subsection{Focal Box}

\begin{figure}[h]%
	\centering
	\includegraphics[width=0.45\textwidth]{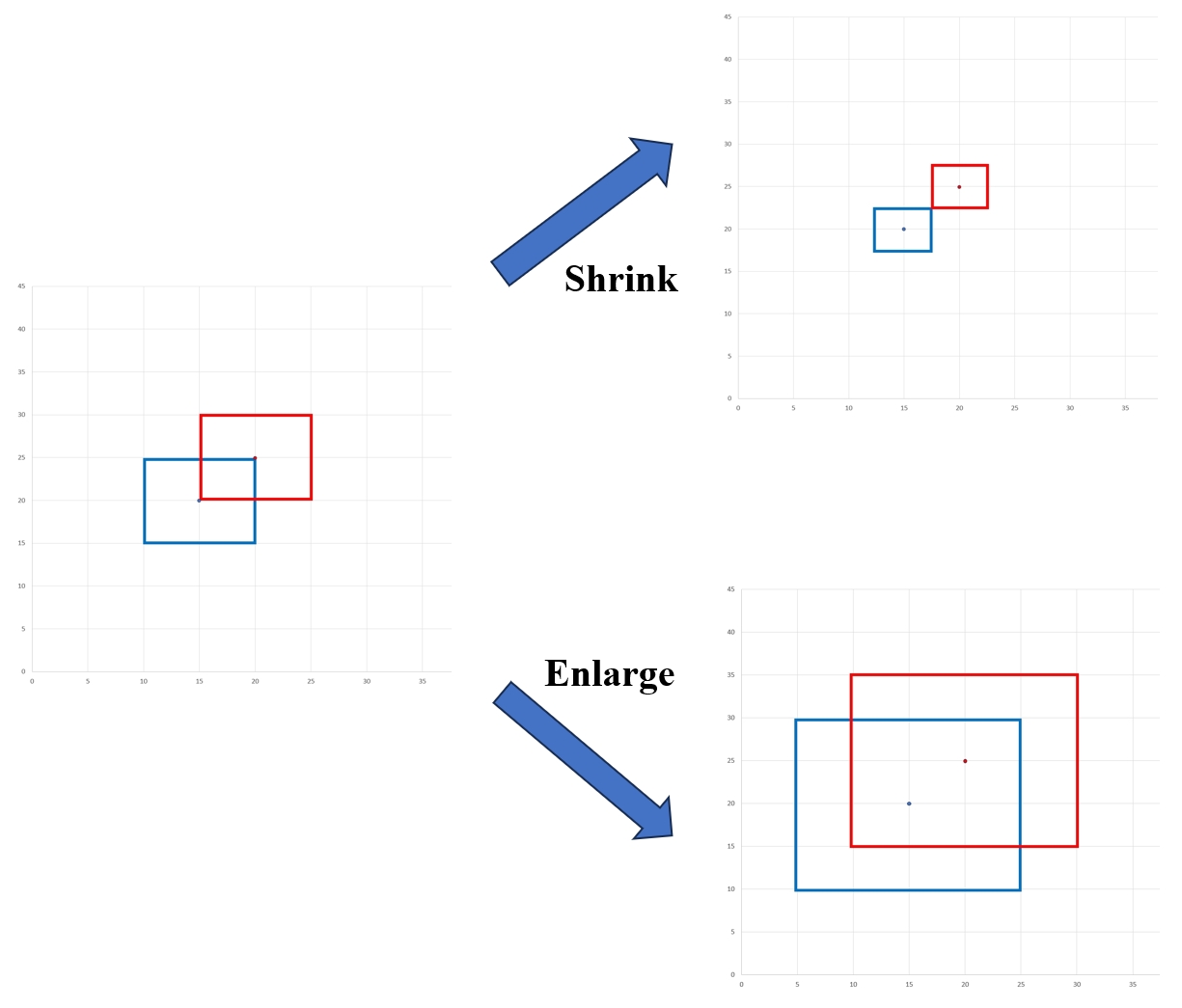}
	\caption{Scaling of prediction boxes and GT boxes (the blue box represents GT and the red box represents prediction)}\label{fig2}
\end{figure}

Instead of assigning different weight to the loss value according to the deviation of the two bounding boxes, we designed a method to enlarge or shrink the prediction box and the GT box, which also achieved the purpose of assigning different weights to different quality prediction boxes. The advantage of this method is that there is no need to carry out any redundant calculation on the bounding box. After obtaining the height and width of the bounding box and the coordinates of the center point, the height and width can be enlarged or shrank in a certain proportion, as shown in Figure~\ref{fig2}.

For a more intuitive sense of the IoU changes caused by scaling, we assume that each cell has a length of 5, and the initial IoU value of the two bounding boxes is:%
\begin{equation}
	\resizebox{.91\linewidth}{!}{$
		\displaystyle
		IoU_{ori} = \frac{inter}{union} = \frac{5\times5}{10\times10+10\times10-5\times5} = 0.143
		$}.
\end{equation}%

The IoU after the bounding box is shrunk by one time is:%
\begin{equation}
	\resizebox{.81\linewidth}{!}{$
		\displaystyle
		IoU_{shr} = \frac{inter}{union} = \frac{0\times0}{5\times5+5\times5-0\times0} = 0
		$}.
\end{equation}%

After the bounding box is enlarged twice, the value of IoU is:%
\begin{equation}
	\resizebox{.91\linewidth}{!}{$
		\displaystyle
		IoU_{enl} = \frac{inter}{union} = \frac{15\times15}{20\times20+20\times20-15\times15} = 0.391
		$}.
\end{equation}%

It can be seen that within a certain range, when shrinking the bounding box, its IoU value will decrease and the calculated box loss will increase. When you zoom in on the bounding box, its IoU value will increase and the calculated box loss will decrease. 

\begin{figure}[h]%
	\centering
	\includegraphics[width=0.45\textwidth]{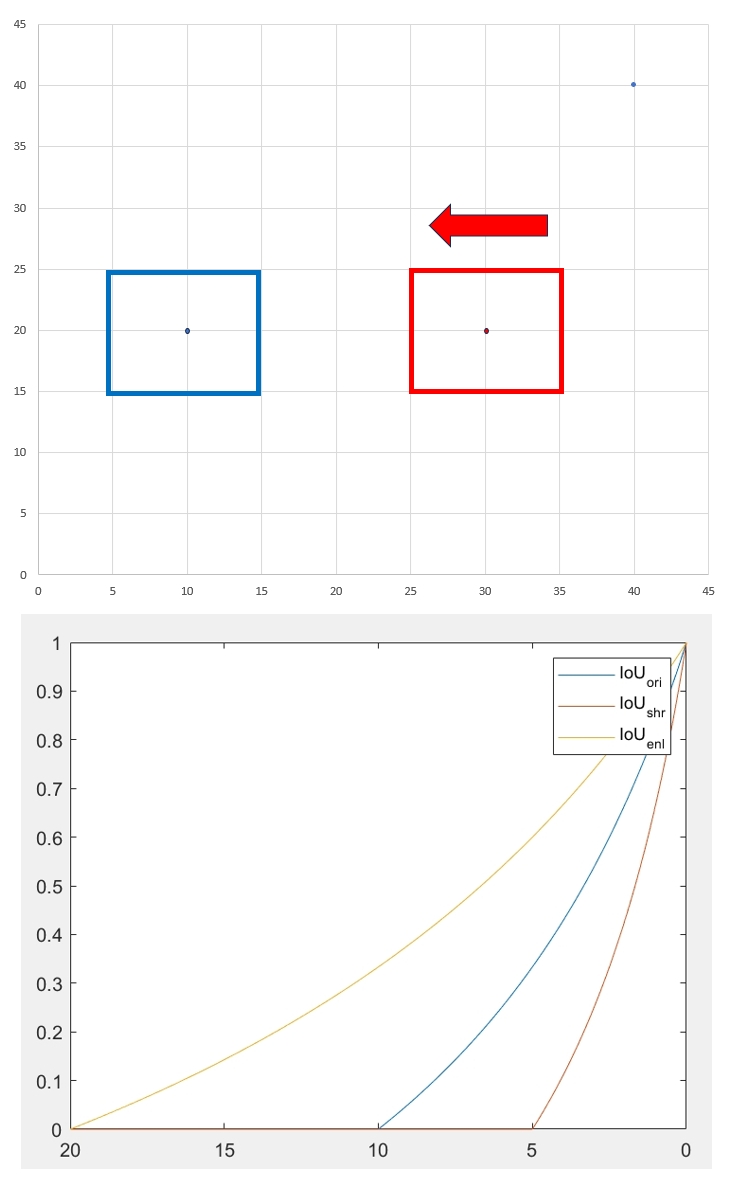}
	\caption{Change of IoU value with different scaling (prediction box from right to left close to GT box)}\label{fig3}
\end{figure}

This change of IoU is also related to the value of IoU itself. As shown in Figure~\ref{fig3}, we draw the change curve of $IoU_{ori}$, $IoU_{shr}$, $IoU_{enl}$ during the process of center distance changing from 20 to 0.

It is not difficult to see from Figure~\ref{fig3} that for those high-quality prediction boxes with large IoU, bounding box reduction can reduce the IoU value and amplify the box regression loss. For those low-quality prediction boxes whose IoU approaches 0, bounding box reduction has little effect on the IoU value. Zooming in on a bounding box has the opposite effect. It is important to note that such a method is not intended to measure the deviation of the prediction box and the GT box more comprehensively, but to amplify the box regression loss of the prediction box and thus obtain more weight.

\subsection{Annealing Strategy Based on Bounding Box Attention}

In analogy to the cosine annealing strategy~\cite{bib14} used for the learning rate during training, we also use this idea for scaling the bounding box. This is because at the beginning of training, there are few high-quality prediction boxes, and only increasing the weight of these few prediction boxes has little effect, but will reduce the convergence speed. Therefore, we should pay more attention to those low-quality prediction boxes, which is in line with popular belief. However, such a way can only achieve the effect of faster convergence, in the case of higher box regression accuracy demand, only focus on low quality prediction box is not advisable. Especially in the dense scenario, there is a large amount of overlap or occlusion between the objects to be detected. If the box regression loss is still trained by many low-quality prediction boxes at this time, there will be a large deviation in the actual detection, which is easy to lead to missing or false detection.

In the previous subsection, we have seen that by shrinking the bounding box, we can increase the regression loss of the box, which is equivalent to increasing the weight of the high-quality prediction box in the current training process; Similarly, by enlarging the bounding box, we can also increase the weight of the low-quality prediction box to achieve faster convergence. Therefore, we design an annealing strategy based on bounding box attention: The dynamic hyperparameter "ratio" was defined as the scaling multiple of the bounding box. At the early stage of training, we adopted the method of enlarging the bounding box and focused the model's attention on the low-quality prediction box to make it converge faster ($ratio\textgreater1$). In the training process, the "ratio" is gradually reduced, and the model's attention is gradually turned to the high-quality prediction box ($ratio\textless1$).

\subsection{Focal Loss}

Different from our focus, Focal Loss~\cite{bib7} also assigns different weights to the loss function. Focal Loss is more willing to pay attention to samples that are difficult to detect, that is, samples with low confidence, rather than the IoU value. We adopted this idea and made further optimizations. Since Focal Loss is proposed for binary classification tasks, in the face of multi-classification problems, especially in the use of YOLO, we simplified this idea and directly used the difference between 1 and the confidence of the prediction box as the weight factor to multiply the calculated IoU. Prediction boxes with less confidence will gain more weight, and in this way we can make the model focus on examples that are harder to reason about.

\subsection{Unified-IoU}

We design a new loss function Unified-IoU (UIoU) by combining the above methods with YOLO's existing box regression loss function. This loss function not only considers the geometric relationship between the prediction box and the GT box, but also takes into account the IoU weight and confidence information, making full use of the known information, so we call it Unified-IoU (UIoU). In particular, we retain the code of box regression loss in the original YOLO, including GIoU,DIoU,CIoU, etc., which can simply switch the calculation method of the loss function, and it is convenient for subsequent researchers to carry out comparative experiments or further improve experiments.

\section{Experiments}

\subsection{Datasets}

To verify the validity of our proposed new loss function, we conducted experiments on two commonly used datasets: VOC2007~\cite{bib15} and COCO2017~\cite{bib16}. In addition, in order to verify the significant improvement in the detection quality of the prediction box, we also conducted tests on the dense dataset CityPersons~\cite{bib17}. We also looked at mAP (mean Average Precision) values at higher IoU thresholds, not just mAP50.

\subsection{Design of The Hyperparameter "ratio"}

\begin{figure}[h]%
	\centering
	\includegraphics[width=0.45\textwidth]{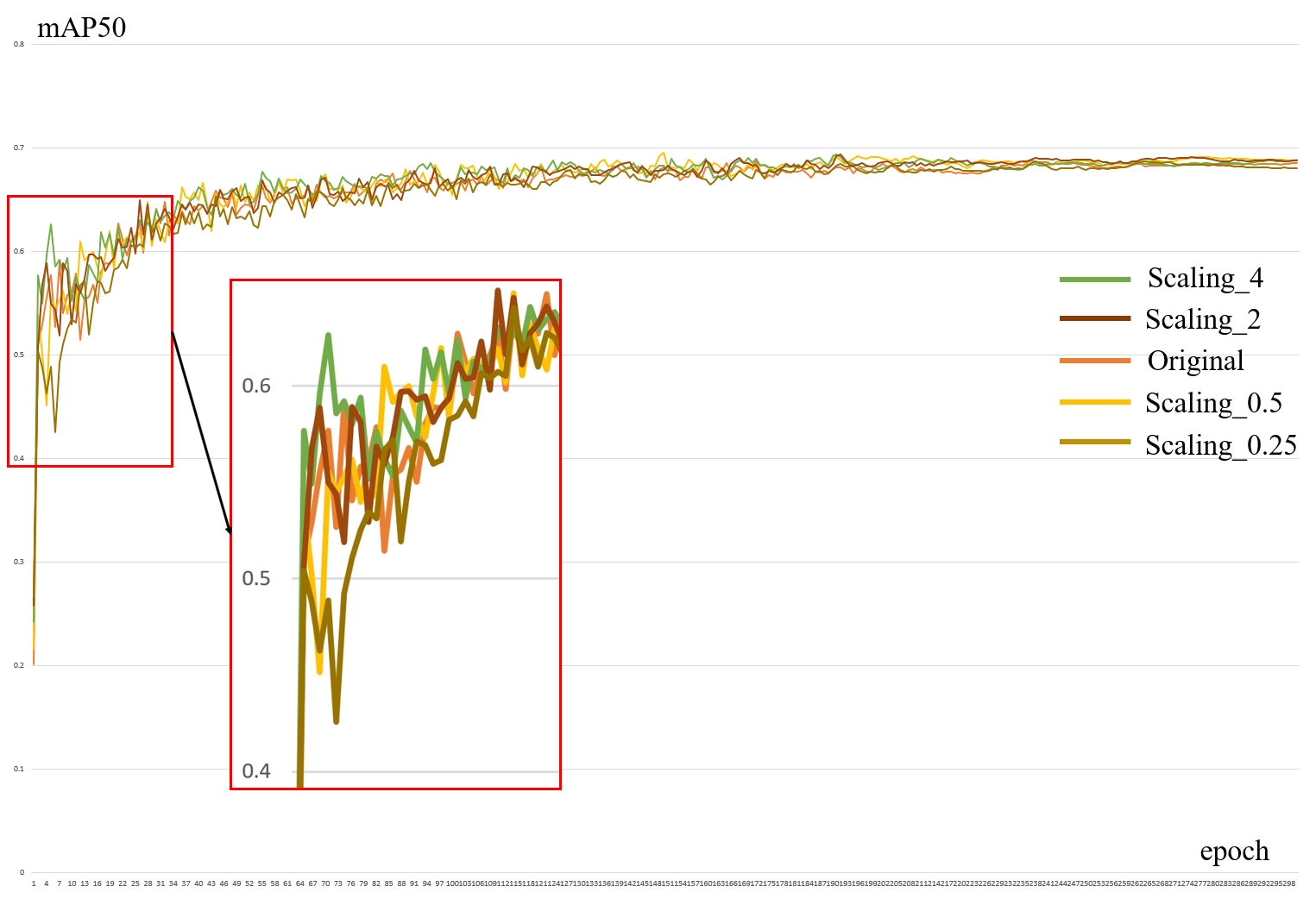}
	\caption{Variation of mAP50 with training epochs under different scaling ratios}\label{fig4}
\end{figure}

If the model focuses on high-quality prediction frames at the beginning, due to the small number of them, the model will converge slowly, which will affect the detection accuracy under certain training rounds. As shown in Figure~\ref{fig4}, we discovered this problem while training on the VOC2007 dataset. "Scaling\_4" represents that in calculating the loss, we enlarge the bounding box by a factor of four. "Original" means that we are scaling the bounding box by a factor of one, that is, doing nothing. When we enlarge the bounding box, the model pays more attention to the large number of low-quality prediction boxes, and the model converges faster and achieves better mAP50 through fewer rounds. After stabilization, the model should gradually turn its attention to high-quality prediction boxes to train more accurate box regression matrices and improve the detection quality of objects.

As mentioned in the previous section, we designed the hyperparameter "ratio" to dynamically adjust the model's attention to different quality prediction boxes, and the value of "ratio" is related to the current training round. We set the initial value of the hyperparameter "ratio" to be 2, so as to achieve a certain precision as soon as possible in the early stage of training, and set the end value of "ratio" to be 0.5, so as to pay more attention to high-quality prediction box in the later stage of training and improve the final object detection quality.

As for the reduction mode of the hyperparameter "ratio", we make reference to the cosine annealing strategy of the learning rate and formulate three reduction modes. Assuming that the training rounds are 300, the relationship between them and the training rounds is as follows:

Linear descent strategy:%
\begin{equation}
	\resizebox{.51\linewidth}{!}{$
		\displaystyle
		ratio = -0.005\times epoch + 2
		$}.
\end{equation}%

Cosine descent strategy:%
\begin{equation}
	\resizebox{.71\linewidth}{!}{$
		\displaystyle
		ratio = 0.75\times cos(\frac{\pi\times epoch}{300}) + 1.25
		$}.
\end{equation}%

Fractional descent strategy:%
\begin{equation}
	\resizebox{.41\linewidth}{!}{$
		\displaystyle
		ratio = \frac{200}{epoch + 100}
		$}.
\end{equation}%

We conducted experiments on these three strategies in the follow-up experiment, each has its advantages and disadvantages, and also showed these three strategies in the code for the convenience of follow-up research.

\subsection{Experiments on VOC2007 and COCO2017 Datasets}

\subsubsection{Result of VOC2007} 

We conducted validation experiments and rich comparison tests on the VOC2007 dataset, and used mAP under different IoU thresholds as an evaluation index to characterize higher-quality object detection. As shown in Table~\ref{VOC2007}, the model training epochs were all 300 for different IoU loss functions. We take the CIoU loss function widely used by YOLO as the baseline and calculate the proportion of the rise point of our proposed UIoU loss relative to the CIoU (baseline). mAP50-75 (the average of 6 mAP across different IoU thresholds) = (mAP50 + mAP55 + $\ldots$ + mAP75) / 6.

\begin{table*}
	\centering
	\begin{tabular}{llllllll}
		\hline
		IoU Loss  & mAP50 & mAP55 & mAP60 & mAP65 & mAP70 & mAP75 & mAP50-75 \\
		\hline
		IoU & 69.1 & 67.4 & 65.1 & 62.1 & 57.5 & 50.9 & 62.02\\
		GIoU & 68.8 & 67.3 & 65.4 & 62.4 & 57.8 & 51.3 & 62.17\\
		DIoU & 68.8 & 66.9 & 65.1 & 62.6 & 57.9 & 51.8 & 62.18\\
		CIoU & 68.8 & 67.0 & 64.8 & 61.6 & 57.6 & 51.3 & 61.85\\
		IoU($\alpha$=3) & 68.6 & 66.8 & 64.8 & 61.3 & 56.4 & 49.6 & 61.25\\
		GIoU($\alpha$=3) & 68.9 & 67.3 & 64.9 & 61.7 & 56.8 & 50.4 & 61.67\\
		DIoU($\alpha$=3) & 68.7 & 66.7 & 64.0 & 60.4 & 56.4 & 50.5 & 61.12\\
		CIoU($\alpha$=3) & 67.4 & 65.6 & 63.8 & 60.3 & 56.2 & 50.4 & 60.62\\
		EIoU & 68.9 & 67.3 & 65.3 & 62.0 & 57.9 & 51.7 & 62.18\\
		SIoU & 68.9 & 67.4 & 65.4 & 62.4 & 57.4 & 50.8 & 62.05\\
		WIoU & 68.7 & 66.8 & 64.6 & 61.6 & 58.1 & 51.6 & 61.90\\
		UIoU(linear) & 69.8 & \textbf{68.3} & \textbf{66.3} & \textbf{63.3} & 58.3 & 51.7 & \textbf{62.95}\\
		& - & +1.94\% & +2.31\% & +2.76\% & - & - & +1.78\% \\
		UIoU(cos) & 69.8 & 68.1 & 66.1 & 62.9 & \textbf{58.5} & \textbf{52.3} & \textbf{62.95}\\
		& - & - & - & - & +1.56\% & +1.95\% & +1.78\% \\
		UIoU(fraction) & \textbf{70.0} & 68.1 & 66.1 & 63.0 & 58.3 & 51.9 & 62.90\\
		& +1.74\% & - & - & - & - & - & - \\
		\hline
	\end{tabular}
	\caption{Experiments with different IoU losses on the VOC2007 dataset, the bolded values represent the maximum values in this column, below it is the proportion of the maximum that increases relative to the CIoU (baseline). $\alpha$=3 means we used $\alpha$-IoU and take $\alpha$=3. linear, cos, fraction represent the different attenuation strategies we use for the hyperparameter "ratio".}
	\label{VOC2007}
\end{table*}

The experimental results show the superiority of our proposed method, which can greatly improve the detection accuracy when the IoU threshold is high, which indicates that the bounding box predicted by our algorithm is more accurate and has higher detection quality. In addition, during the training process, we also maintained the accuracy when the IoU threshold was low, and even slightly improved. We found that the attenuation strategy for different hyperparameter "ratio" will also affect the detection accuracy, which may be related to the speed at which the model shifts its attention from low-quality prediction boxes to high-quality prediction boxes. 

\subsubsection{Result of COCO2017} 

We conducted similar experiments on COCO2017 dataset, and the experimental results are shown in Table~\ref{COCO2017}. Our method also achieved great improvement on mAP. Compared with the baseline CIoU loss function, mAP50, mAP75, mAP95, and mAP50-95 increased by 0.2\%, 0.8\%, 0.44\%, and 0.5\%, respectively, with a relative increase of 0.353\%, 2.00\%, 18.88\%, and 1.34\%.

The experimental results show that the new loss function proposed by us can train a more accurate target detection model, and the position regression of the prediction box is more accurate, which is conducive to the detection of objects, especially the high-quality detection of objects.

\begin{table}
	\centering
	\scalebox{0.85}{
	\begin{tabular}{lllll}
		\hline
		IoU Loss  & mAP50 & mAP75 & mAP95 & mAP50-95\\
		\hline
		CIoU & 56.7 & 40.1 & 2.33 & 37.3 \\
		\hline
		\hline
		UIoU(linear) & \textbf{56.9} & \textbf{40.9} & 2.49 & \textbf{37.8} \\
		Relative improv.\% & +0.353\% & +2.00\% & - & +1.34\% \\
		\hline
		UIoU(cos) & 56.6 & 40.1 & 2.74 & 37.5 \\
		Relative improv.\% & - & - & - & - \\
		\hline
		UIoU(fraction) & 56.8 & 40.8 & \textbf{2.77} & 37.7 \\
		Relative improv.\% & - & - & +18.88\% & - \\
		\hline
	\end{tabular}}
	\caption{Comparison of the proposed method (UIoU) with baseline (CIoU) for 300 epochs of training on COCO2017 dataset.}
	\label{COCO2017}
\end{table}

\subsection{Experiments on CityPersons Datasets}

For high-quality object detection, we conducted additional experiments on the CityPersons dataset. The CityPersons dataset is a subset of CityScape~\cite{bib18}, with 2,975 images for training, 500 and 1,575 images for validation and testing. The average number of pedestrians in an image is 7, and there are a large number of objects blocking each other. For this kind of dense dataset, if the prediction box offset is large, it is easy to affect the model's detection of another nearby target, resulting in missing detection. Therefore, the dataset has very high requirements on the quality of the prediction box, and can also verify the validity of the loss function proposed by us.

As shown in Table~\ref{CityPersons}, our UIoU Loss does not perform well in the CityPersons dataset, which we think may be Focal Loss affecting our model. There is a certain relationship between the confidence level of the model and the quality of the prediction box during the training process. When we pay attention to difficult examples with low confidence level, it also pays more attention to low-quality prediction boxes, which are no longer applicable to dense pedestrian datasets like CityPersons. Therefore, for the training of this dataset, we improved Focal Loss and applied the idea of Focal Loss in reverse, called "Focal-inv", and conducted a confirmatory experiment on this idea.

\begin{table*}
	\centering
	\scalebox{0.8}{
	\begin{tabular}{llllllllllll}
		\hline
		IoU Loss  & AP50 & AP55 & AP60 & AP65 & AP70 & AP75 & AP80 & AP85 & AP90 & AP95 & AP50-95 \\
		\hline
		CIoU & 62.2 & 59.4 & 55.6 & 52.0 & 46.4 & 39.9 & 30.6 & 20.3 & 8.62 & 0.989 & 37.5\\
		EIoU & 62.8 & 59.6 & 56.5 & 52.3 & 46.5 & 39.9 & 30.6 & 20.7 & 8.49 & 1.03 & 37.8\\
		SIoU & 61.8 & 58.6 & 55.6 & 51.4 & 45.8 & 39.1 & 30.1 & 19.8 & 8.06 & 0.909 & 37.1\\
		WIoU & 61.4 & 58.7 & 55.4 & 50.7 & 44.9 & 38.2 & 30.1 & 19.9 & 7.31 & 0.564 & 36.7\\
		\hline
		UIoU(linear) $\downarrow$ & 59.2 & 56.0 & 52.7 & 48.2 & 42.5 & 35.4 & 26.0 & 15.6 & 5.71 & 0.417 & 34.2\\
		& - & - & - & - & - & - & - & - & - & - & -\\
		UIoU(cos) $\downarrow$ & 60.0 & 56.9 & 53.0 & 47.9 & 42.1 & 34.0 & 24.6 & 14.6 & 4.89 & 0.272 & 33.8\\
		& - & - & - & - & - & -14.79\% & -19.61\% & -28.08\% & -43.62\% & -72.50\% & -\\
		UIoU(fraction) $\downarrow$ & 58.5 & 55.2 & 51.6 & 47.0 & 41.0 & 34.7 & 25.5 & 15.1 & 5.80 & 0.315 & 33.5\\
		& -5.95\% & -7.07\% & -7.19\% & -9.62\% & -11.64\% & - & - & - & - & - & -10.67\%\\
		\hline
		UIoU-Focal-inv(linear) $\uparrow$ & 63.3 & 60.4 & \textbf{57.5} & 53.1 & 47.0 & \textbf{40.5} & 31.6 & 20.8 & 9.40 & 1.18 & 38.5\\
		& - & - & +3.42\% & - & - & +1.50\% & - & - & - & - & -\\
		UIoU-Focal-inv(cos) $\uparrow$ & \textbf{63.4} & \textbf{60.7} & 57.4 & \textbf{53.6} & \textbf{47.6} & 39.8 & 30.7 & 19.9 & 9.75 & 1.23 & 38.4\\
		& +1.93\% & +2.19\% & - & +3.08\% & +2.59\% & - & - & - & - & - & -\\
		UIoU-Focal-inv(fraction) $\uparrow$ & 63.1 & 60.4 & 57.1 & 52.8 & 47.5 & 40.0 & \textbf{32.1} & \textbf{21.8} & \textbf{9.89} & \textbf{1.27} & \textbf{38.6}\\
		& - & - & - & - & - & - & +4.90\% & +7.39\% & +14.73\% & +28.41\% & +2.93\%\\
		\hline
	\end{tabular}}
	\caption{The effects of different IoU loss functions on the CityPersons dataset, with the addition of Focal-inv related experiments.}
	\label{CityPersons}
\end{table*}

\begin{figure*}[h]%
	\centering
	\includegraphics[width=1\textwidth]{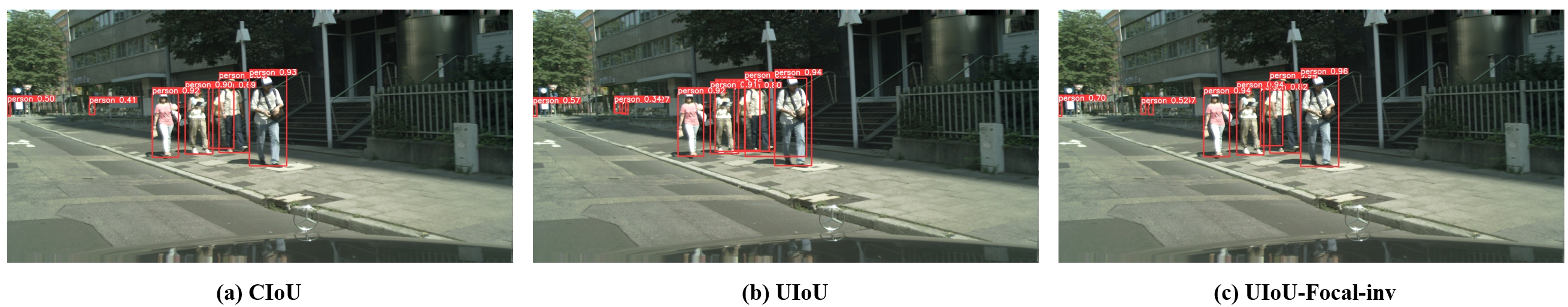}
	\caption{Detection results of models trained with different IoU loss functions on the CityPersons dataset}\label{fig5}
\end{figure*}

Table~\ref{CityPersons} illustrates that the problems faced by dense datasets are different from conventional datasets. Dense datasets are more focused on the training of high-quality prediction boxes, and the model should pay attention to simple prediction boxes with large IoU values. As shown in Table~\ref{CityPersons}, the larger the IoU threshold, the more obvious the improvement effect relative to CIoU (baseline), and the model can detect objects with higher quality.  Figure~\ref{fig5} shows the detection effect of the training model under the CityPersons dataset with different IoU loss functions. As shown in Figure~\ref{fig5}(b), the model trained with the original "Focal Loss" idea was greatly affected by the gradient of low-quality samples, and many extra wrong prediction boxes were generated in the detection. This problem would become prominent in the dense datasets, resulting in a sharp decline in detection accuracy. In Figure~\ref{fig5}(c), we corrected the point of "focusing on the difficult to predict box" and adopted "Focal-inv", which greatly improves the accuracy of bounding box regression and confidence.

\subsection{Ablation Experiment}

In order to further validate the above viewpoint on Focal Loss in a dense dataset and to clarify the effect of other improvements on the model during training, we also conducted ablation experiments on the CityPersons dataset. As shown in Table~\ref{Ablation}, we studied four different situations in which only Focal Box was added, Focal Box with hyperparameter "ratio" attenuation strategy, Focal Loss only and Focal-inv only were added. "Only adding Focal Box" means that the reduction ratio of the bounding box is kept constant at 0.5 during the training process, and the model always keeps the focus on high-quality prediction box. Focal-inv is the opposite idea of Focal Loss, which focuses on objects that are easier to detect.

\begin{table}
	\centering
	\scalebox{0.75}{
		\begin{tabular}{llllll}
			\hline
			Baseline  & Focal Box & \quad\quad"ratio" & Focal Loss & Focal-inv & AP90\\
			          &           & attenuation strategy &            &      & (\%)    \\
			\hline
			\checkmark &  &  &  &  & 8.62\\
			\checkmark & \checkmark &  &  &  & 9.61\\
			\checkmark & \checkmark & \checkmark &  &  & 9.14\\
			\checkmark &  &  & \checkmark &  & 6.28\\
			\checkmark &  &  &  & \checkmark & 8.83\\
			\checkmark & \checkmark & \checkmark &  & \checkmark & \textbf{9.89}\\
			\hline
	\end{tabular}}
	\caption{The ablation experiment results on the CityPersons dataset, in order to demonstrate our improvement in high-quality detection, adopted AP90 as the evaluation index.}
	\label{Ablation}
\end{table}

The experimental results in Table~\ref{Ablation} show the advantages of our improvements and verify our views: scaling at different proportions can change the model's attention to prediction boxes of different qualities, and reducing the bounding box will help the model pay more attention to high-quality prediction boxes and improve the high-quality detection performance of the model; Using the hyperparameter "ratio" to dynamically adjust the scaling ratio can make the model converge faster, but it will slightly affect the high-quality detection effect. The idea of Focal Loss is no longer suitable for dense datasets. Instead, dense datasets should focus on objects that are easier to detect, because there is a correlation between confidence and IoU loss. Our improved Focal-inv shows more superior high-quality detection results.

\section{Conclusion}

In this paper, we propose a new method to assign weights to different quality prediction boxes. Different from the previous ones, the weights we assign are dynamically related to the position, size and other coefficients of the current prediction box and Ground Truth (GT) box, rather than simply multiplying by a certain coefficient. During training, we used the hyperparameter "ratio" to adjust the model's attention to prediction boxes with different qualities, so as to achieve a balance between the training speed and the high quality detection of objects. Finally, we combined the idea of "Focal Loss" and verified the superiority of the improved "Focal-inv" through experiments on dense datasets. Taking all the above into account, we creatively proposed a Unified-IoU (UIoU), and conducted comparative experiments on two commonly used datasets, VOC2007 and COCO2017, which proved the validity of our proposed UIoU.

In the future, we will explore more efficient box regression loss functions because of their widespread use and importance in deep-learning tasks. In addition, as a method for high-quality object detection, we will demonstrate the effect of our UIoU in more dense datasets.

\bibliographystyle{named}
\bibliography{ijcai24}

\end{document}